\documentclass[review]{elsarticle}

\usepackage{lineno,hyperref}
\usepackage{amsmath, amssymb, amsfonts, wasysym, tipa, textcomp, amsbsy, graphics,subcaption, graphicx, setspace, fancyhdr, MnSymbol, mathrsfs , color,placeins, lipsum,xfrac,amsthm,listings, multirow}
\usepackage{booktabs}
\usepackage[utf8]{inputenc}
\usepackage[english]{babel}
\usepackage[labelfont=bf]{caption}
\usepackage{enumerate}
\definecolor{mygreen}{RGB}{28,172,0} 
\definecolor{mylilas}{RGB}{170,55,241}

\usepackage{float}
\restylefloat{table} 
\usepackage{tkz-graph}
\usetikzlibrary{arrows}
\usepackage{pinlabel}
\usepackage{float}
\floatstyle{plaintop}
\restylefloat{table}

\modulolinenumbers[250]

\begin{document}

\begin{frontmatter}

\title{Deep learning for video game genre classification}

\author{Yuhang Jiang, Lukun Zheng} 
\address{1906 College Heights Blvd, Bowling Green, KY 42101}

\author[mymainaddress]{Western Kentucky University} 

\cortext[mycorrespondingauthor]{Corresponding Author: Lukun Zheng}
\ead{lukun.zheng@wku.edu}


\begin{abstract}
Video game covers and textual descriptions are usually the very first impression to its consumers and they often convey important information about the video games. Video game genre classification based on its cover and textual description would be utterly beneficial to many modern identification, collocation, and retrieval systems. At the same time, it is also an extremely challenging task due to the following reasons: First, there exists a wide variety of video game genres, many of which are not concretely defined. Second, video game covers vary in many different ways such as colors, styles, textual information, etc, even for games of the same genre. Third, cover designs and textual descriptions may vary due to many external factors such as country, culture, target reader populations, etc. With the growing competitiveness in the video game industry, the cover designers and typographers push the cover designs to its limit in the hope of attracting sales. The computer-based automatic video game genre classification systems become a particularly exciting research topic in recent years. In this paper, we propose a multi-modal deep learning framework to solve this problem. The contribution of this paper is four-fold. First, we compiles a large dataset consisting of 50,000 video games from 21 genres made of cover images, description text, and title text and the genre information. Second, image-based and text-based, state-of-the-art models are evaluated thoroughly for the task of genre classification for video games. Third, we developed an efficient  and salable multi-modal framework based on both images and texts.  Fourth, a thorough analysis of the experimental results is given and future works to improve the performance is suggested. The results show that the multi-modal framework outperforms the current state-of-the-art image-based or text-based models. Several challenges are outlined for this task. More efforts and resources are needed for this classification task in order to reach a satisfactory level.
\end{abstract}


\end{frontmatter}

\linenumbers

\section{Introduction}
Genres are stylistic categories of literature, music, or other forms of art or entertainments characterized by their forms, contents, subject matters, and the like. Genres are often used to identify, retrieve, and organize works of interest. Genre classification is the process of grouping works together based on certain stylistic similarities. It has been used in a wide range of areas such as music, paintings, film, books, etc. These important tasks were traditionally done manually, which has a lot of limitations on cost, time and other resources. With the growing capacity of computational powers of modern computers, many different automatic genre classification techniques have been developed and used in many domains such as movies, books, paintings, etc. In this study, we focus on the task of genre classification of video games based on the cover and textual description. To the best of our knowledge, this is the first attempt on automatic genre classification using  a deep learning approach. 

Videos games have been one of the most popular, profitable, and influential forms of entertainment across the world. There have been many controversies surrounding video games. Many studies show that video games can be very helpful in education across different age groups and comprehension levels \cite{S2003,M2009,B2017,L2019,K2020}. However, Others point out that video games can cause many problems in terms of health, time wasting, and, even, violence crimes \cite{BNVF2017,F2018,G2020}. Despite of the massive debates and arguments in academia and real life, it is persuasive that the video game market has been exploding across the world over the years. 

Genre and its classification systems play a significant role in the development of video games. Even though the genre classification for video games is similar to that for other forms of media, it has its own specialties and challenges. First, unlike other media such as books and movie, video games are generally interactive with its player \cite{CLC2017}. Second, there are constant formation and conceptualization of new genres over time due to the fast development of new video games. Third, existing genres may shift over time as developers, players and the media come up with new names. Finally, video games have a short history compared with other forms of media \cite{N2013}. 

Recently, deep learning approaches have reached high performances across a wide variety of problems \cite{RWDBALY2016, GHV2017, SWS2017,CLTSSX2018, KP2018,MC2019,SGM2019,ZCZ2020,GWHLLB2020}. In particular, some deep convolutional neural networks can achieve a satisfactory level of performance on many visual recognition and categorization tasks, exceeding human performances. One of the most attractive qualities of these techniques is that they can perform well without any external hand-designed resources or task-specific feature engineering.  The theoretical foundations of deep learning are well rooted in the classical neural network (NN) literature. It involves many hidden neurons and layers as an architectural advantage in addition to the input and output layers \cite{RWDBALY2016}. A deep convolutional neural network is universal, meaning that it can be used to approximate any continuous function to an arbitrary accuracy when the depth of the neural network is large enough \cite{Z2020}. In this paper, we aim to develop three deep learning algorithm for the task of video game genre classification: image-based approach using the game covers, text-based approach using the textual descriptions, and multimodal approach using both the game covers and textual description.

\begin{figure}[t]
		\centering
		\begin{subfigure}[t]{28mm}
    		\includegraphics[width=28mm, height=40mm]{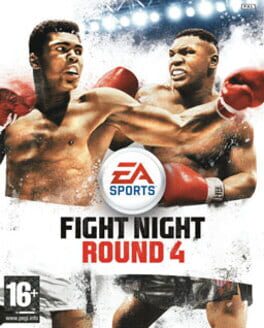}
    		\caption{}
		\end{subfigure}
		\begin{subfigure}[t]{28mm}
		    \includegraphics[width=28mm, height=40mm]{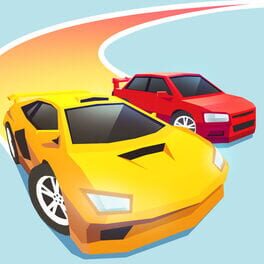}
		    \caption{}
		\end{subfigure}
		\begin{subfigure}[t]{28mm}
		    \includegraphics[width=28mm, height=40mm]{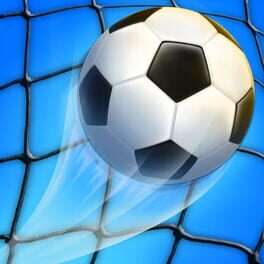}
		    \caption{}
		\end{subfigure}
		\begin{subfigure}[t]{28mm}
		    \includegraphics[width=28mm, height=40mm]{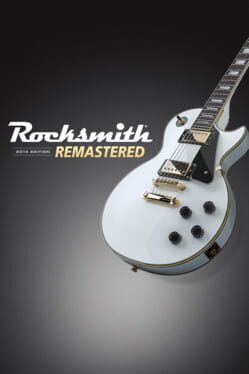}
		    \caption{}
		\end{subfigure}
		\caption{ Cover images conveying important information about the genres of corresponding video games. The genres are: (a) \emph{Fighting;}
		(b) \emph{Racing}; (c) \emph{Sports}; and (d) \emph{Music}.                       
		}
		\label{fig1:covers}
	\end{figure}

Covers are usually the very first impression to its consumers and they often convey important information about the video games. Figure \ref{fig1:covers} presents some sample video game covers. For instance, in Figure 1(a), the cover image features two boxers which indicates that it is a fighting game. The two cars side-by-side on the road in Figure 1 (b) tells that it is a racing game. The soccer ball in Figure \ref{fig1:covers} (c) indicates that it is a sport game. Lastly, in Figure \ref{fig1:covers} (d), there is a guitar in the cover, which implies that this is a music game. 

In addition to the cover images, we also use the game descriptions for genre classification.  Game descriptions are designed to suitably express the key concepts and mechanics inherent in video games. The main purpose of game description is to express the objects involved in a games and the set of rules that induce transitions, resulting in a state-action space \cite{ELLSTT2013}. It usually contains important information about the its genre. Consider the following example of a description for the game of \emph{BOUNCING BALLS}:\\
\emph{"Use your mouse to aim and shoot. Destroy the balls by shooting them into groups of three or more. Clear all of the balls to get to the next level. The faster you clear the board the more points you'll get! The game ends when a ball hits the bottom."}
In this description, the keywords "aim", "shoot", "shooting", etc indicate that it is a shooter game. Hence we will also explore the use of the game descriptions for the task of genre classification. 

Finally, we consider a multimodal deep learning architecture based on both the game cover and description for the task of genre classification. This approach involves two steps: (1) A neural network is trained on the classification task for each modality. (2) Intermediate representations are extracted from each network and combined in a multimodal learning step. We believe that relating information from both modalities will help to improve the performance in genre classification for video games.  

The main contributions of this study includes:
\begin{enumerate}[(1)]
\item We created a large dataset of 50,000 video games including game cover images, description text, title text, and genre information. This dataset can be used for a variety of studies such as text recognition from images, automatic topic mining, and so on. This dataset will be made available to the public in the future. 
\item State-of-the-art models are evaluated thoroughly for the task of video game genre classification. These include five image-based models and two text-based models using deep transfer learning methods utilizing the information from existing pre-trained models. 
\item  We developed a multi-modal deep learning archtecture based on two modalities: one for images and the other for texts. The information from both modalities are then combined using concatenation method. We believe that the mergence of information from both modalities would help increase the classification accuracy rate for this task. 
\item A comprehensive analysis of the difficulties in this video game genre classification task is provided and possible suggestions are made for future work to improve the performance. 
\end{enumerate}

The rest of the paper is structured as follows. Section 2 presents related works about book cover classification. Section 3 elaborates on the details of the models. In section 4, we discuss the experimental results. The last section concludes the paper and discusses future work. 
\section{Related Works}
In recent years, automated genre classification has drawn a wide range of attention from different domains by leveraging the strength of the deep neural network. 

In book genre classification, Chiang et al. \cite{CGW2015} implemented transfer learning with convolutional neural networks on the cover image along with natural language processing on the title text. A data set consisting of 6000 book covers from five genres obtained from OpenLibrary.org were utilized for their study. Iwana et al. \cite{Iwa2016} attempted to conduct book genre classification using only the visual clues provided by its cover. To solve this task, they created a large dataset consisting of 57,000 samples from 30 genres and adapted AlexNet \cite{KSH2012} pre-trained on ImageNet \cite{DDSLLF2009}.  Buczkowski et al. \cite{BSK2018} created another dataset consisting of 160k book covers crawled from GoodReads.com from over 500 genres, from which they picked the top 13 most popular genres and grouped all the remaining books under a 14th class called ``Others". In \cite{BRVS2019}, the authors utilized a multinomial logistic regression model to classify book genres based on extracted image features and title features. Their approach consists of three stages: image feature extraction using the Xception model \cite{C2017}, title feature extraction using the GloVe Model \cite{PSM2014}, and classification based on the combined extracted features.  Kundu and Zheng \cite{KZ2020} proposed a multimodal deep learning framework based on book covers and text directly extracted from the book covers, which resembles the way people obtain both visual and textual information from the cover. They provided a detailed evaluation of the state-of-the-art models for the task of book genre classification.

In music genre classification, traditional methods often focus on feature engineering that extracts features using domain knowledge which are then used in the designed learning algorithm such as support vector machine and k-nearest neighbors \cite{L2000,TC2002,GDPW2004}. More recently, deep learning approaches has been proposed for music genre classification using visual representations of audio as input to convolutional neural networks (CNNs) \cite{DBS2011,ZLXX2016,COS2017,D2018}. Text-based deep learning approaches have also been proposed. Oramas et al. \cite{OEL2016} used a large dataset of about 65k albums constructed from Amazon customer reviews and performed experiments on music genre classification. Fang et al. \cite{FGLW2017} investigate genre classification of music using descourse-based features extracted from lyrics. There have also been multimodal studies on music genre classification in the literature, many of which combine audio and song lyrics \cite{LGH2008,ONBS2017,OBNS2018}. 

However, genre classification for video games has drawn few attentions in academia. To the best knowledge of the authors, there has been no studies on this task using a deep learning approach. Clarke et al. \cite{CLC2017} explores the affordances and limitations of video game genre classification from a library and information science perspective, discussing various purposes of genre relating to video games. Amiriparian et al. \cite{ACGPOS2019} proposed an audio-based video game genre classification model using a linear support vector vector machine classifier. They created a new corpus collected entirely from social multimedia and extracted three different feature representations from the game audio files. In this study, we explore video genre classification using three different deep learning approaches: one based on game cover image, one based on game description text, and one based on both game cover image and game description text.

\section{Methodology}
\subsection{Data Set Preparation}

In this study, we compiled a large dataset consisting of 50,000 video games with their cover images, description text, title text, and genre information from from IGDB.com, a video game database, intended for both game consumers and video game professionals alike. Most of the data available are user-generated, including the genre information. There were in total 21 genres found in the original dataset. These original genres were labeled without considering the internal relationships among some of the genres. For instance,  \emph{Real Time Strategy} is a sub-genre of the genre \emph{Strategy}; \emph{Hack
and slash/Beat'em up} is a sub-genre of the genre \emph{Fighting}. In such cases, we grouped genres like these as one genre. After these modification, we have in total 15 different genres as shown in Table \ref{tab: genre table}.

\begin{table}[ht]
\centering
\caption{The 15 genres of video games in our dataset}
\def\arraystretch{0.8}
\begin{tabular}{lllll}
\hline\hline \\[-0.1ex]
Adventure &Arcade &Fighting &Indie &Music\\[1.2ex]
Pinball& Platform& Puzzle& Quiz/Trivia&Racing\\[1.2ex]
Role-Playing&Shooter&Simulator&Sport&Strategy\\[1.2ex]\hline\hline 
\end{tabular}
\label{tab: genre table}
\end{table}

Another problem encountered is that many video games have two or more genres. In such cases, we randomly selected one genre for each video game and assume that each video game has a single genre. Doing so simplifies our problem to single-label classification. Downloaded cover images have different sizes and ratios. All the cover images were re-sized to 224 px by 224 px so that they have the same dimension.  The description texts were encoded using a standard tokenization in which each word in the text is tokenized and then assigned a unique integer for representation. In this study we only encode words that appear at least 10 times in the whole dataset to avoid noises caused by insignificant words for classification. We first introduce some baseline image-based and text-based models and then a multimodal learning architectures proposed in this paper. We split the dataset into three parts: 70\% training samples, 10\% validation samples, and 20\% test samples.  

\subsection{Image-Based and Text-Based Models}
\paragraph{Image-based models} We use several image-based models as our baseline models:  MobileNet-V1 \cite{HZCKWWA2017},, MobileNet-V2 \cite{SHZZC2018}, Inception-V1 \cite{SLJSRAR2015}, Inception-V2 \cite{SLVA2017},  and ResNet-50\cite{HSS2018}. These models were pre-trained on large datasets and achieved satisfactory performance on other similar image classification tasks. In this paper, we utilize these pre-trained models through transfer learning technique using the pre-trained values of the parameters involved in the models.

\paragraph{Text-based models} In addition to these image-based models, we also evaluated two text-based models for this task. One is recurrent neural networks (RNN) with Long Short-Term Memory (LSTM) \cite{HS1997}. It allows the network to accumulate past information and thus be able to learn the long-term dependencies which are very common in textual data. The other one is Universal Sentence Encoder \cite{CYKHL2018}. It is a pre-trained sentence embedding and designed to leverages the encoder from Transformer which is a multi-attention head that helps the model ``attend" to the relevant information.

\subsection{The Multi-Modal Model with Simple Concatenation}
\begin{figure}[ht]
\centering
\includegraphics[height=7 cm,width=12 cm]{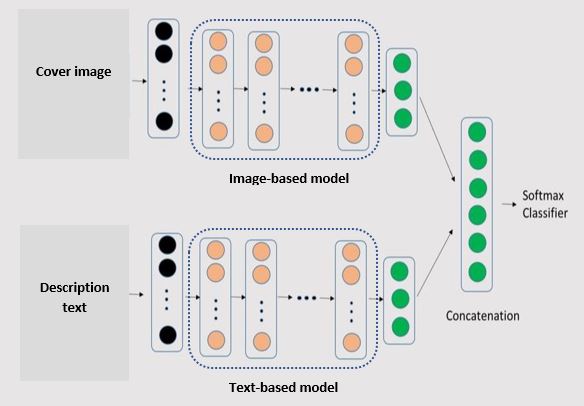}
\caption{The proposed multi-modal model with simple concatenation at a higher layer including deep networks (input layer, hidden layers, and output layer), concatenation layer and classifier softmax.}
\label{fig2:simple concatenation}
\end{figure}

The proposed multi-modal model includes a visual modality and a textual modality with simple concatenation at a higher layer, which is shown in Figure \ref{fig2:simple concatenation}. 
The model contains three parts: deep networks (input layer, hidden layers, and output layer), feature concatenation and softmax classifier. The concatenation occurs at a higher layer instead of the input layer since concatenation at the input layer often causes 1) intractable training effort; 2) over-fitting due to pre-maturely learned features from both modalities; and 3) failure to learn implicit associations between modalities with different underlying features \cite{TYJ2018}. This model first learns the two modalities separately with two different flows, and then concatenate their features at a higher layer. It proves to be effective when two modalities have different basic representations, such as the visual data and the textual data. We use the sparse categorical-cross-entropy loss to train the neural network.

\section{Experimental Results}
\subsection{Implementation}
\begin{itemize}
\item[(1)] \emph{Image-based models}. The image-based models mentioned in the previous section, we utilized the pre-trained models on ImageNet dataset and added a flatten layer of 1024 hidden units, and an output layer of 30 units on top of every pre-trained models. The softmax activation function was used for the output layer. We used sparse-categorical-cross-entropy loss with Adam optimizer with a default learning rate of 0.001 for all these models. 
\item[(2)] \emph{Text-based models}. We used a simple vanilla LSTM with 256 memory units for the text-based RNN model. We used the pre-trained Universal Sentence Encoder model \cite{CYKHL2018}. The softmax activation function was used for the output layer. We used sparse-categorical-cross-entropy loss with Adam optimizer with a default learning rate of 0.001 in both models.. 
\item[(3)] \emph{Multi-modal models}. We chose the best image-based model (ResNet-50) and use it in the visual modality and chose the best text-based model (Universal Sentence Encoder) and use it in the textual modality. For the multi-modal model with simple concatenation, we freeze the chosen pre-trained models and extract features using the ReLU activation function. Then we concatenate these features from both modalities and feed them into a final fully connected layer with a softmax activation function. 
\item[(4)] \emph{Experimental setup}. All the experiments are executed with Python in Google Colab using a Nvidia Tesla K80 GPU. We use Keras (an open-source deep learning library) with the TensorFlow backend. 
\end{itemize}
\subsection{Results}
Table \ref{tab: accuracy table} shows the classification accuracy comparison among the models including  the five image-based models, two text-based models, and the multi-modal model on the test set. Among the image-based models, ResNet-50 achieved the best performance with a top-1 accuracy of 31.4\% and a top-3 accuracy of 61.7\%. The two text-based models had much better results. The RNN-LSTM model achieved a top-1 accuracy of 44.3\% and a top-3 accuracy of 72.1\%. The Universal Sentence Encoder model achieved a top-1 accuracy of 47.7\% and a top-3 accuracy of 76.3. Finally, the multi-modal model had the best performance. The simple concatenation model achieved a top-1 accuracy of 49.9\% and a top-3 accuracy of 79.9\%.

\begin{table}[ht]
\centering
\caption{Accuracy comparison of the models including the image-based models, text-based models, and multi-modal models on the test set}
\begin{tabular}{p{0.58\textwidth}llp{0.25\textwidth}lp{0.25\textwidth}}
\hline\hline 
&\multicolumn{2}{c}{Accuracy}\\[0.3ex]\cmidrule{2-3} 
Models&Top 1(\%)  &Top 3 (\%)\\  [0.5ex] \hline
\emph{Image-based models}&&\\[1.2ex]
MobileNet-V1 &29.2 &58.7\\
MobileNet-V2 &28.7 &57.1\\
Inception-V1 &27.4 &54.4\\
Inception-V2 &28.3 &56.5\\
\textbf{ResNet-50} &\textbf{31.4} &\textbf{61.7} \\ [0.5ex]\hline
\emph{Text-based models}&&\\[1.2ex]
RNN-LSTM &44.3 &72.1\\
\textbf{Universal Sentence Encoder} &\textbf{47.7} &\textbf{76.3}\\ [0.5ex]\hline
\emph{Multi-modal model} &&  \\[1.2ex]
\textbf{Simple concatenation} &\textbf{49.9} &\textbf{79.9}\\\hline\hline 
\end{tabular}
\label{tab: accuracy table}
\end{table}

Table \ref{tab: genre accuracy comparison} presents the classification accuracy comparison on the test data among the best image-based model(ResNet-50), the best text-based model (Universal Sentence Encoder), and the multi-modal model for individual genres. There are several observations from this table. First, it can be seen that the multi-modal model outperforms the other two models in most of the genres. Second, the multi-modal model achieved high accuracy rates for some of the genres such as \emph{Sport} and \emph{Racing}, while, in some other genres such as \emph{Arcade} and \emph{Quiz/Trivia}, its accuracy rates are very low. Third, the image-based model have the lowest accuracy rates for all genres except \emph{Quiz/Trivia}, which indicates that genre classification based on only cover images is a very challenging problem for video games. Results suggest that the combination of modalities outperforms single modality approaches.

\begin{table}[H]
\centering
\caption{Top 1 accuracy comparison (in percentages) among the best image-based model (ResNet-50), the best text-based model (USE model) and the multi-modal model for individual genres on the test se. Here USE represents Universal Sentence Encoder.}
\def\arraystretch{0.8}
\begin{tabular}{lccc}
\hline\hline 
Genre &ResNet-50  &USE model &Multi-modal model\\  [0.5ex]
\hline 
Adventure &39.5&63.0& 54.1   \\
Arcade    &2.0 &2.6  & 7.7  \\
Fighting  &35.6& 52.7 & 51.7   \\
Indie     &41.7&30.8  & 46.7  \\
Music     &8.6 &54.7 & 67.2  \\
Pinball   &28.1 &56.3 & 56.3   \\
Platform  &16.3& 30.0 & 39.2  \\
Puzzle    &30.2 & 43.8 & 47.2   \\
Quiz/Trivia&6.3 & 1.6 & 7.81   \\
Racing     &54.9& 66.4& 69.5  \\
Role Playing&12.2 &38.0 & 50.5   \\
Shooter     &35.6 &58.3  & 56.5   \\
Simulator   &15.5& 35.7 & 38.0 \\
Sport       &58.6 &71.9 & 71.9  \\
Strategy    &18.0 & 52.4& 49.6 \\\hline\hline
\end{tabular}
\label{tab: genre accuracy comparison}
\end{table}

\noindent Finally, there exists a certain degree of consistency among these three models among these genres. That is, if one model has a large accuracy rate in one genre, then the other two models tend to have large accuracy rates on this genre as well; if one model has a small accuracy rate in one genre, then the other two models tend to have small accuracy rates on this genre as well. For instance, these three models all have relatively high accuracy rates on \emph{Sport}, while, for \emph{Arcade}, they all have very low accuracy rates.

\begin{figure}[ht]
\centering
\includegraphics[scale=.4]{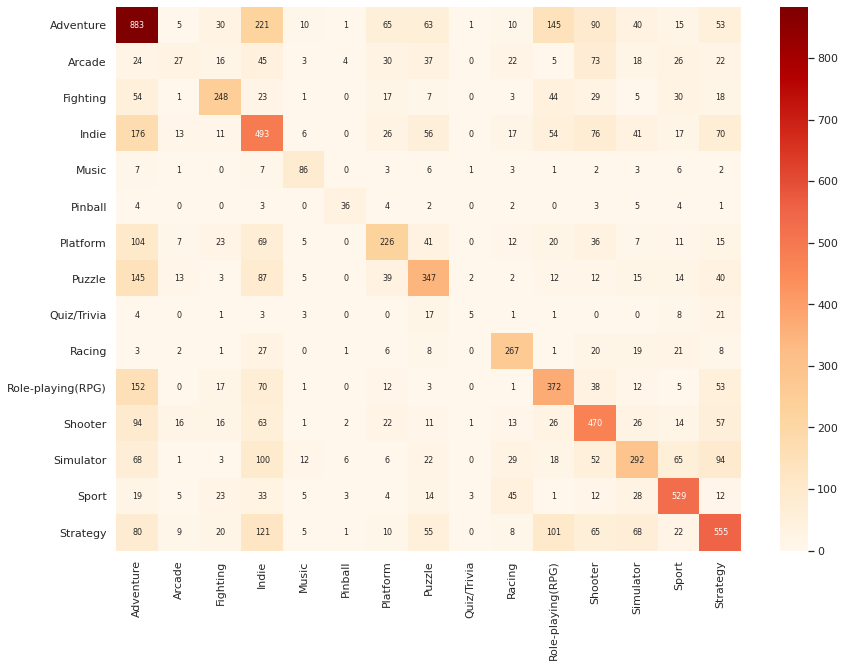}
\caption{Confusion matrix of the multi-modal model on the test data. The horizontal axis represents the predicted genres and the vertical axis represents the observed genres.}
\label{fig: confusion-matrix}
\end{figure}

Figure \ref{fig: confusion-matrix} presents the confusion matrix of the multi-modal model on the test data, with the horizontal axis representing the predicted genres and the vertical axis representing the observed genres. 
The diagonal entries represent the number of video games of each genre in the test set that were classified correctly by the multi-modal model. For instance, the entry ``248" at the (3,3) position of the matrix tells that 248 video games from \emph{Fighting} are classified correctly as \emph{Fighting}, resulting in a top-1 accuracy of 248/480=51.7\%, where 480 is the total number of video games with the genre \emph{Fighting} from the test set. The off-diagonal entries provide detailed information about the numbers of misclassifications among each genre. For instance, the entry ``3" in the (6, 4) position of the matrix tells that 3 video games in \emph{Pinball} were misclassified as \emph{Indie}.
One observation is that the diagonal entries are dominant for most of the genres, which implies that our classification model correctly classifies the video games to their actual genre for most of the times. Another observation is that misclassification often occurs for closely related genres. For instance, 95 video games in \emph{Shooter} were misclassified as \emph{Adventure}, since these two genres are closely related.

\subsection{Discussion}
Even though the proposed multi-modal model has a relatively good performance for most of the individual genres as shown in Table \ref{tab: genre accuracy comparison}, their performances in some of the genres such as \emph{Arcade} and \emph{Quiz/Trivia} are far from satisfactory. In this section, we summarize the challenges present in the task of video game genre classification. 

\begin{figure}[ht]
		\centering
		\begin{subfigure}[t]{28mm}
		    \includegraphics[width=28mm, height=40mm]{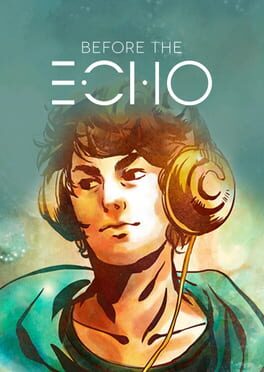}
		    \caption{}
		\end{subfigure}
		\begin{subfigure}[t]{28mm}
		    \includegraphics[width=28mm, height=40mm]{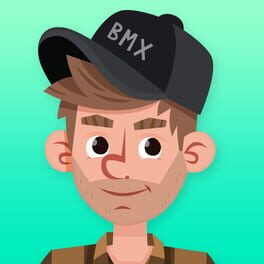}
		    \caption{}
		\end{subfigure}
		\begin{subfigure}[t]{28mm}
    		\includegraphics[width=28mm, height=40mm]{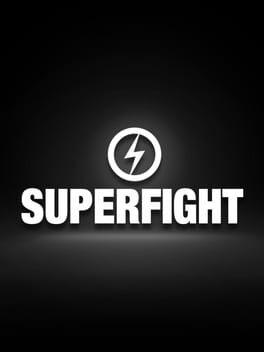}
    		\caption{}
		\end{subfigure}
		\begin{subfigure}[t]{28mm}
		    \includegraphics[width=28mm, height=40mm]{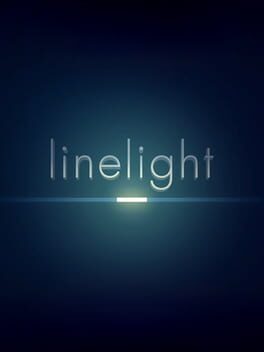}
		    \caption{}
		\end{subfigure}
\caption{Low inter-genre variance. (a) genre: \emph{Music}; (b) genre: \emph{Sport}; (c) genre: \emph{Indie}; (d) genre:  \emph{Puzzle}. 
		}
		\label{fig5:low-inter-variance}
	\end{figure}
First, there exist many cover images that look similar but belong to different genres \cite{LSSRIUA2020}. Low inter-genre variance refers to the fact that cover images of different genres look very similar. For instance, Figures \ref{fig5:low-inter-variance} (a) and \ref{fig5:low-inter-variance} (b) are very similar to each other but they belong to different genres. Similarly, Figures \ref{fig5:low-inter-variance} (c) and \ref{fig5:low-inter-variance} (d) are very similar to each other but they belong to different genres as well. 

\begin{figure}[ht]
		\centering
		\begin{subfigure}[t]{28mm}
    		\includegraphics[width=28mm, height=40mm]{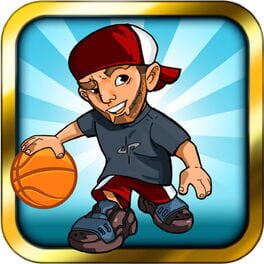}
    		\caption{}
		\end{subfigure}
		\begin{subfigure}[t]{28mm}
		    \includegraphics[width=28mm, height=40mm]{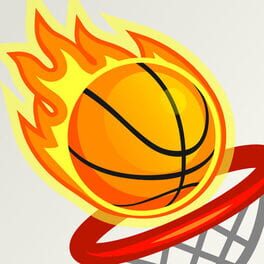}
		    \caption{}
		\end{subfigure}
		\begin{subfigure}[t]{28mm}
		    \includegraphics[width=28mm, height=40mm]{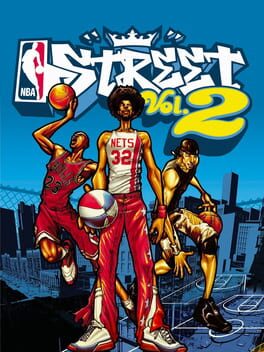}
		    \caption{}
		\end{subfigure}
		\begin{subfigure}[t]{28mm}
		    \includegraphics[width=28mm, height=40mm]{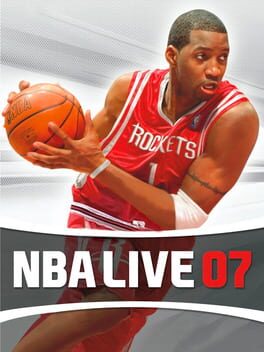}
		    \caption{}
		\end{subfigure}		
		\caption{High intra-genre variance. All of these video games are about basketball game, belonging to the same genre \emph{Sport} but their covers are very different.  
		}
		
		\label{fig6:high-intra-variance}
	\end{figure}

Second, video game cover images of the same genre often exhibit big difference among them. For instance, all the video games in Figure \ref{fig6:high-intra-variance} are about basketball games but their art styles are very different from each other. 
 
	Another factor that makes it difficult for prediction in this task is that there exist many misleading cover images which may convey inconsistent information regarding the genres of the video games. 
\begin{figure}[ht]
		\centering
		\begin{subfigure}[t]{28mm}
    		\includegraphics[width=28mm, height=40mm]{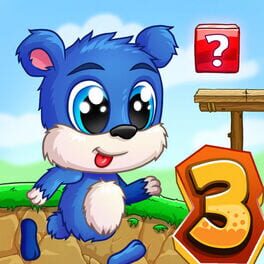}
    		\caption{}
		\end{subfigure}
		\begin{subfigure}[t]{28mm}
		    \includegraphics[width=28mm, height=40mm]{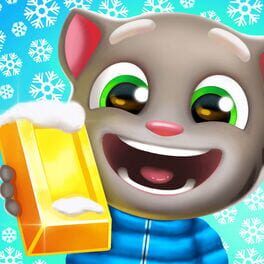}
		    \caption{}
		\end{subfigure}
		\begin{subfigure}[t]{28mm}
		    \includegraphics[width=28mm, height=40mm]{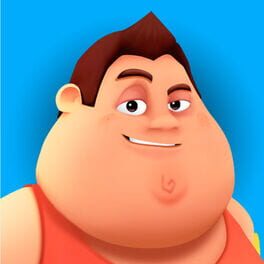}
		    \caption{}
		\end{subfigure}
		\begin{subfigure}[t]{28mm}
		    \includegraphics[width=28mm, height=40mm]{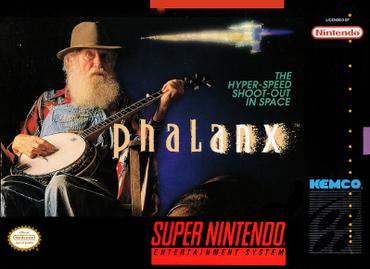}
		    \caption{}
		\end{subfigure}
		\caption{Misleading cover images about the genre. (a), (b), and (c) are of the genre:\emph{Sport}; (d) is of the genre \emph{Shooter}.
		}
		\label{fig7:inconsistant-tx-image}
	\end{figure}		
In Figure \ref{fig7:inconsistant-tx-image} (a),(b), and (c), there are little information from these cover images that implies its true genre \emph{Sport}. In (d), the cover image include an old man holding a musical instrument, from which one may think that it is of the genre \emph{Music}. However, it actual genre is \emph{Shooter}.

Finally, video games involve an interactive element that inherent in these games \cite{CLC2017}. This makes the genre classification problems more complicated than other forms of works such as books, music, and movies.

\section{Conclusion}
In this paper, we proposed three deep learning approaches: one based on cover images, one based on description text, and one multimodal model based on both cover image and description text, for the task of video game genre classification. We created a large dataset of 50,000 video games including game cover images, description text, title text, and genre information. This dataset can be used for a variety of studies such as text recognition from images, automatic topic mining, and so on and will be made available to the public in the future.  In addition, we evaluated several state-of-the-art image-based models and text-based models. We also developed a multimodal model by concatenating features from a image modality and a text modality. Generally speaking, text-based models perform better in general than image-based models and the the multi-modal model outperforms all other models.  We also outlined several challenges in the task of genre classification for video games. In order to solve this task to a better level of performance, more efforts are needed on the creation of better data sets and the development of more sophisticated models. 

\vspace{.2 cm}  



\begin{thebibliography}{99}

\bibitem{ACGPOS2019} Amiriparian, S., Cummins, N., Gerczuk, M., Pugachevskiy, S., Ottl, S., \& Schuller, B. (2019). “are you playing a shooter again?!” deep representation learning for audio-based video game genre recognition. IEEE Transactions on Games.
\bibitem{B2017}Barr, M. (2017). Video games can develop graduate skills in higher education students: A randomised trial. Computers \& Education, 113, 86-97.

\bibitem{BNVF2017} Bean, A. M., Nielsen, R. K., Van Rooij, A. J., \& Ferguson, C. J. (2017). Video game addiction: The push to pathologize video games. Professional Psychology: Research and Practice, 48(5), 378.

\bibitem{BRVS2019} Biradar, G. R., Raagini, J. M., Varier, A., \& Sudhir, M. (2019). Classification of Book Genres using Book Cover and Title. In 2019 IEEE International Conference on Intelligent Systems and Green Technology (ICISGT) (pp. 72-723). IEEE.

\bibitem{BSK2018} Buczkowski, P., Sobkowicz, A., \& Kozlowski, M. (2018). Deep Learning Approaches towards Book Covers Classification. In ICPRAM (pp. 309-316).

\bibitem{CLTSSX2018} Cao, C., Liu, F., Tan, H., Song, D., Shu, W., Li, W., ... \& Xie, Z. (2018). Deep learning and its applications in biomedicine. Genomics, proteomics \& bioinformatics, 16(1), 17-32.

\bibitem{CYKHL2018} Cer, D., Yang, Y., Kong, S. Y., Hua, N., Limtiaco, N., John, R. S., ... \& Sung, Y. H. (2018). Universal sentence encoder. arXiv preprint arXiv:1803.11175.

\bibitem{CGW2015} Chiang, H., Ge, Y., \& Wu, C. (2015). Classification of Book Genres By Cover and Title.

\bibitem{C2017} Chollet, F. (2017). Xception: Deep learning with depthwise separable convolutions. In Proceedings of the IEEE conference on computer vision and pattern recognition (pp. 1251-1258).

\bibitem{CLC2017} Clarke, R. I., Lee, J. H., \& Clark, N. (2017). Why video game genres fail: A classificatory analysis. Games and Culture, 12(5), 445-465.


\bibitem{COS2017} Costa, Y. M., Oliveira, L. S., \& Silla Jr, C. N. (2017). An evaluation of convolutional neural networks for music classification using spectrograms. Applied soft computing, 52, 28-38.

\bibitem{DDSLLF2009} Deng, J., Dong, W., Socher, R., Li, L. J., Li, K., \& Fei-Fei, L. (2009). Imagenet: A large-scale hierarchical image database. In 2009 IEEE conference on computer vision and pattern recognition (pp. 248-255). Ieee.


\bibitem{DBS2011} Dieleman, S., Brakel, P., \& Schrauwen, B. (2011). Audio-based music classification with a pretrained convolutional network. In 12th International Society for Music Information Retrieval Conference (ISMIR-2011) (pp. 669-674). University of Miami.

\bibitem{D2018} Dong, M. (2018). Convolutional neural network achieves human-level accuracy in music genre classification. arXiv preprint arXiv:1802.09697.

\bibitem{ELLSTT2013} Ebner, M., Levine, J., Lucas, S. M., Schaul, T., Thompson, T., \& Togelius, J. (2013). Towards a video game description language. Schloss Dagstuhl-Leibniz-Zentrum fuer Informatik.
\bibitem{FGLW2017} Fang, J., Grunberg, D., Litman, D. T., \& Wang, Y. (2017). Discourse Analysis of Lyric and Lyric-Based Classification of Music. In ISMIR (pp. 464-471).
\bibitem{F2018}Ferguson, C. J. (2018). Violent Video Games, Sexist Video Games, and the Law: Why Can't We Find Effects?. Annual Review of Law and Social Science, 14, 411-426.

\bibitem{GHV2017} Goh, G. B., Hodas, N. O., \& Vishnu, A. (2017). Deep learning for computational chemistry. Journal of computational chemistry, 38(16), 1291-1307.
\bibitem{GDPW2004}Gouyon, F., Dixon, S., Pampalk, E., \& Widmer, G. (2004, June). Evaluating rhythmic descriptors for musical genre classification. In Proceedings of the AES 25th International Conference (pp. 196-204).

\bibitem{G2020} Guggisberg, M. (2020). Sexually explicit video games and online pornography-the promotion of sexual violence: a critical commentary. Aggression and violent behavior, 53, e101432-e101432.
\bibitem{GWHLLB2020} Guo, Y., Wang, H., Hu, Q., Liu, H., Liu, L., \& Bennamoun, M. (2020). Deep learning for 3d point clouds: A survey. IEEE transactions on pattern analysis and machine intelligence.
\bibitem{HS1997} Hochreiter, S., \& Schmidhuber, J. (1997). Long short-term memory. Neural computation, 9(8), 1735-1780.

\bibitem{HZCKWWA2017}Howard, A. G., Zhu, M., Chen, B., Kalenichenko, D., Wang, W., Weyand, T., ... \& Adam, H. (2017). Mobilenets: Efficient convolutional neural networks for mobile vision applications. arXiv preprint arXiv:1704.04861.

\bibitem{HSS2018} Hu, J., Shen, L., \& Sun, G. (2018). Squeeze-and-excitation networks. In Proceedings of the IEEE conference on computer vision and pattern recognition (pp. 7132-7141).
\bibitem{Iwa2016} Iwana, B. K., Rizvi, S. T. R., Ahmed, S., Dengel, A., \& Uchida, S. (2016). Judging a book by its cover. arXiv preprint arXiv:1610.09204.
\bibitem{KP2018} Kamilaris, A., \& Prenafeta-Boldú, F. X. (2018). Deep learning in agriculture: A survey. Computers and electronics in agriculture, 147, 70-90.

\bibitem{K2020}Keeler, K. R. (2020). Video Games in Music Education: The Impact of Video Games on Rhythmic Performance. Visions of Research in Music Education, (37).

\bibitem{KSH2012} Krizhevsky, A., Sutskever, I., \& Hinton, G. E. (2012). Imagenet classification with deep convolutional neural networks. In Advances in neural information processing systems (pp. 1097-1105).

\bibitem{KZ2020} Kundu, C., \& Zheng, L. (2020). Deep multi-modal networks for book genre classification based on its cover. arXiv preprint arXiv:2011.07658.
\bibitem{LGH2008} Laurier, C., Grivolla, J., \& Herrera, P. (2008, December). Multimodal music mood classification using audio and lyrics. In 2008 Seventh International Conference on Machine Learning and Applications (pp. 688-693). IEEE.

\bibitem{L2000} Logan, B. (2000, October). Mel frequency cepstral coefficients for music modeling. In Ismir (Vol. 270, pp. 1-11).
\bibitem{L2019}Lougheed, T. (2019). Video games bring new aspects to medical education and training.

\bibitem{LSSRIUA2020}Lucieri, A., Sabir, H., Siddiqui, S. A., Rizvi, S. T. R., Iwana, B. K., Uchida, S., ... \& Ahmed, S. (2020). Benchmarking Deep Learning Models for Classification of Book Covers. SN Computer Science, 1, 1-16.


\bibitem{MC2019} Mater, A. C., \& Coote, M. L. (2019). Deep learning in chemistry. Journal of chemical information and modeling, 59(6), 2545-2559.

\bibitem{M2009}Mayo, M. J. (2009). Video games: A route to large-scale STEM education?. Science, 323(5910), 79-82.
\bibitem{N2013} Newman, J. A. (2013). Videogames. Routledge.

\bibitem{OBNS2018} Oramas, S., Barbieri, F., Nieto, O., \& Serra, X. (2018). Multimodal deep learning for music genre classification. Transactions of the International Society for Music Information Retrieval. 2018; 1 (1): 4-21.
\bibitem{OEL2016} Oramas, S., Espinosa-Anke, L., \& Lawlor, A. (2016, August). Exploring customer reviews for music genre classification and evolutionary studies. In The 17th International Society for Music Information Retrieval Conference (ISMIR 2016), New York City, United States of America, 7-11 August 2016.
\bibitem{ONBS2017} Oramas, S., Nieto, O., Barbieri, F., \& Serra, X. (2017). Multi-label music genre classification from audio, text, and images using deep features. arXiv preprint arXiv:1707.04916.
\bibitem{PSM2014}Pennington, J., Socher, R., \& Manning, C. D. (2014). Glove: Global vectors for word representation. In Proceedings of the 2014 conference on empirical methods in natural language processing (EMNLP) (pp. 1532-1543).




\bibitem{RWDBALY2016} Ravì, D., Wong, C., Deligianni, F., Berthelot, M., Andreu-Perez, J., Lo, B., \& Yang, G. Z. (2016). Deep learning for health informatics. IEEE journal of biomedical and health informatics, 21(1), 4-21.
\bibitem{SHZZC2018} Sandler, M., Howard, A., Zhu, M., Zhmoginov, A., \& Chen, L. C. (2018). Mobilenetv2: Inverted residuals and linear bottlenecks. In Proceedings of the IEEE conference on computer vision and pattern recognition (pp. 4510-4520).
\bibitem{SWS2017} Shen, D., Wu, G., \& Suk, H. I. (2017). Deep learning in medical image analysis. Annual review of biomedical engineering, 19, 221-248.
\bibitem{S2003} Squire, K. (2003). Video games in education. Int. J. Intell. Games \& Simulation, 2(1), 49-62.
\bibitem{SGM2019} Strubell, E., Ganesh, A., \& McCallum, A. (2019). Energy and policy considerations for deep learning in NLP. arXiv preprint arXiv:1906.02243.
\bibitem{SLJSRAR2015} Szegedy, C., Liu, W., Jia, Y., Sermanet, P., Reed, S., Anguelov, D., ... \& Rabinovich, A. (2015). Going deeper with convolutions. In Proceedings of the IEEE conference on computer vision and pattern recognition (pp. 1-9).

\bibitem{SLVA2017} Szegedy, C., Ioffe, S., Vanhoucke, V., \& Alemi, A. A. (2017, February). Inception-v4, inception-resnet and the impact of residual connections on learning. In Thirty-first AAAI conference on artificial intelligence.


\bibitem{TYJ2018} Tang, L., Yang, Z. X., \& Jia, K. (2018). Canonical correlation analysis regularization: an effective deep multiview learning baseline for RGB-D object recognition. IEEE Transactions on Cognitive and Developmental Systems, 11(1), 107-118.

\bibitem{TC2002} Tzanetakis, G., \& Cook, P. (2002). Musical genre classification of audio signals. IEEE Transactions on speech and audio processing, 10(5), 293-302.

\bibitem{ZLXX2016} Zhang, W., Lei, W., Xu, X., \& Xing, X. (2016, September). Improved Music Genre Classification with Convolutional Neural Networks. In Interspeech (pp. 3304-3308).

\bibitem{ZCZ2020} Zhang, Z., Cui, P., \& Zhu, W. (2020). Deep learning on graphs: A survey. IEEE Transactions on Knowledge and Data Engineering.

\bibitem{Z2020} Zhou, D. X. (2020). Universality of deep convolutional neural networks. Applied and computational harmonic analysis, 48(2), 787-794.

\end{thebibliography}
\end{document}